\documentclass[conference, 10pt]{IEEEtran}
\IEEEoverridecommandlockouts
\usepackage{cite}
\usepackage{amsmath,amssymb,amsfonts}
\usepackage{graphicx}
\usepackage{textcomp}
\usepackage{xcolor}
\usepackage{url}
\usepackage{placeins}
\usepackage{float}
\usepackage{adjustbox}
\usepackage{tabularx,colortbl}
\usepackage{ifthen}
\usepackage{caption,subcaption}
\usepackage{amsmath}
\usepackage{booktabs}
\renewcommand{\thetable}{\Roman{table}}

\usepackage{enumitem}
\usepackage{stfloats}
\usepackage{wrapfig}
\usepackage{verbatim}
\usepackage{booktabs}
\usepackage{flushend}
\usepackage{hyperref}
\usepackage{cleveref}
\usepackage{amsmath,amssymb,amsfonts}
\usepackage{mathtools} 
\usepackage{multirow}
\usepackage{algorithm}
\usepackage{algpseudocode}
\usepackage{pifont}

\def\BibTeX{{\rm B\kern-.05em{\sc i\kern-.025em b}\kern-.08em
    T\kern-.1667em\lower.7ex\hbox{E}\kern-.125emX}}

\begin{document}
\newcolumntype{P}[1]{>{\centering\arraybackslash}p{#1}}
\newcolumntype{M}[1]{>{\centering\arraybackslash}m{#1}}
\setlength{\textfloatsep}{10pt plus 1.0pt minus 2.0pt}
\setlength{\dbltextfloatsep}{10pt plus 1.0pt minus 2.0pt}
\setlength{\floatsep}{5pt plus 1.0pt minus 2.0pt}
\setlength{\dblfloatsep}{5pt plus 1.0pt minus 2.0pt}
\setlength{\intextsep}{10pt plus 1.0pt minus 2.0pt}
\title{INTACT: Inducing Noise Tolerance through Adversarial Curriculum Training for LiDAR-based Safety-Critical Perception and Autonomy}
\author{Nastaran Darabi, Divake Kumar, Sina Tayebati, and Amit Ranjan Trivedi \\
University of Illinois at Chicago (UIC) \\
\texttt{\{ndarab2, amitrt\}@uic.edu}
}
\maketitle

\begin{abstract}
In this work, we present INTACT, a novel two-phase framework designed to enhance the robustness of deep neural networks (DNNs) against noisy LiDAR data in safety-critical perception tasks. INTACT combines \textit{meta-learning} with \textit{adversarial curriculum training} (ACT) to systematically address challenges posed by data corruption and sparsity in 3D point clouds. The meta-learning phase equips a teacher network with task-agnostic priors, enabling it to generate robust saliency maps that identify critical data regions. The ACT phase leverages these saliency maps to progressively expose a student network to increasingly complex noise patterns, ensuring targeted perturbation and improved noise resilience. INTACT’s effectiveness is demonstrated through comprehensive evaluations on object detection, tracking, and classification benchmarks using diverse datasets, including KITTI, Argoverse, and ModelNet40. Results indicate that INTACT improves model robustness by up to 20\% across all tasks, outperforming standard adversarial and curriculum training methods. This framework not only addresses the limitations of conventional training strategies but also offers a scalable and efficient solution for real-world deployment in resource-constrained safety-critical systems. INTACT's principled integration of meta-learning and adversarial training establishes a new paradigm for noise-tolerant 3D perception in safety-critical applications. INTACT improved KITTI Multiple Object Tracking Accuracy (MOTA) by \textbf{9.6\%} (64.1\% $\rightarrow$ 75.1\%) and by \textbf{12.4\%} under Gaussian noise (52.5\% $\rightarrow$ 73.7\%). Similarly, KITTI mean Average Precision (mAP) rose from \textbf{59.8\% to 69.8\%} (50\% point drop) and \textbf{49.3\% to 70.9\%} (Gaussian noise), highlighting the framework’s ability to enhance deep learning model resilience in safety-critical object tracking scenarios.


\end{abstract}

\vspace{2pt}
\textbf{Index Terms:} Adversarial Curriculum Training; LiDAR Point Clouds; Cluster-Based Saliency Maps; Meta-Learning

\section{Introduction}
The integration of artificial intelligence (AI) into safety-critical systems, such as autonomous vehicles and remote sensing platforms, necessitates models that are not only highly accurate but also robust, reliable, and interpretable \cite{kim2018robust}. While deep learning models excel in controlled conditions, their performance can degrade significantly in real-world scenarios due to unexpected environmental perturbations, such as adverse weather, sensor noise, or hardware failures. These challenges are particularly pronounced when operating on complex sensor modalities such as Light Detection and Ranging (LiDAR), which has become increasingly critical for enhanced situational awareness beyond what cameras can achieve \cite{kim2020extended}. As a result, developing AI frameworks that effectively address data and sensing uncertainties while maintaining high performance is critical for ensuring safety and operational reliability.

LiDAR sensors, widely used for their precise 3D mapping capabilities, play a key role in perception for robotics, autonomous systems, and environmental monitoring. LiDAR provides high-resolution 3D spatial information, enabling precise distance measurements and environmental mapping, even in low-light or complete darkness. However, LiDAR data is prone to two dominant forms of degradation: \textit{data loss} and \textit{data corruption}. Data loss arises from occlusions, reflective surfaces, or sensor field-of-view limitations, leading to incomplete point clouds with significant gaps in spatial coverage \cite{nex2014uav, lu2019l3}. On the other hand, data corruption occurs due to multi-path reflections, environmental conditions (e.g., rain, fog, or dust), or sensor vibration, introducing erroneous or jittery points into the data. These degradations affect downstream tasks such as object detection, mapping, and navigation, where even small errors can cascade into critical system failures.

While Deep Neural Networks (DNNs) have demonstrated exceptional performance in processing LiDAR and other sensor modalities, their reliability diminishes significantly in the presence of noisy or corrupted data \cite{park2025lidar, li2020deep}. Standard training paradigms often fail to generalize well to noisy conditions and necessitate more sophisticated approaches to improve robustness. Existing solutions, such as adversarial training \cite{madry2017towards} and curriculum learning \cite{bengio2009curriculum}, offer partial remedies but lack systematic strategies to address the complex and varied noise characteristics inherent in real-world LiDAR data.

To overcome these limitations, we propose a novel framework, INTACT, designed to enhance DNN robustness against LiDAR noise through the integration of \textit{meta-learning} and \textit{adversarial curriculum training} (ACT). In INTACT, meta-learning establishes strong initial representations that generalize effectively across diverse and noisy conditions by leveraging task-agnostic priors \cite{finn2017model}. Additionally, by incorporating cluster-based saliency maps, our framework identifies and preserves critical regions of the data that are most relevant for accurate predictions \cite{yang2018two}. This targeted preservation ensures that noise in less salient regions does not disproportionately affect model performance. Furthermore, ACT systematically introduces noise patterns of increasing complexity during training, emulating real-world degradation scenarios to improve noise resilience \cite{cai2018curriculum, sarkar2021adversarial}. This comprehensive methodology enables DNNs with enhanced noise tolerance, making them particularly suitable for safety-critical applications.

To evaluate INTACT, we conduct experiments across object detection, tracking, and classification using multiple datasets. For object detection, we use the KITTI benchmark \cite{geiger2012we} for detecting vehicles, pedestrians, and cyclists in urban settings, and the Argoverse dataset \cite{chang2019argoverse} to assess performance on additional object classes, such as buses. For classification, we leverage the ModelNet40 dataset, which includes 3D CAD models spanning 40 object categories \cite{wu20153d}. Results show that INTACT-trained models improve robustness by up to 20\% under challenging noise conditions, outperforming conventional adversarial and curriculum training approaches. INTACT improved KITTI Multiple Object Tracking Accuracy (MOTA) by 9.6\% (64.1\% → 75.1\%) and by 12.4\% under Gaussian noise (52.5\% → 73.7\%). Similarly, KITTI mean Average Precision (mAP) increased from 59.8\ to 69.8\% under a 50\% point drop and from 49.3\% to 70.9\% under Gaussian noise, demonstrating INTACT’s ability to enhance model resilience in safety-critical tracking scenarios. Additionally, INTACT improves object detection by mitigating noise and data loss. On the Argoverse dataset, bus detection mAP rose from 20.7\% to 31.0\% (50\% drop) and from 17.4\% to 32.3\% (Gaussian noise). Car detection mAP improved from 57.3\% to 69.3\% (50\% drop) and from 51.6\% to 68.7\% (Gaussian noise), highlighting its effectiveness in improving perception models under challenging conditions.

The remainder of this paper is structured as follows: Sec. II reviews key concepts in LiDAR sensing, their noise characteristics, and point cloud processing. Sec. III details INTACT, including the meta-learning strategy and ACT methodology. Sec. IV presents experimental results across detection, tracking, and classification tasks. Sec. V concludes.

\section{Background}  

\subsection{Energy-Reliability Trade-offs of LiDAR at the Edge}
LiDAR systems deployed at the edge face significant challenges due to limited computational and energy resources, compounded by noise-related issues such as data loss and corruption. These constraints hinder real-time processing, as corrupted point clouds—caused by sensor noise or environmental interference—propagate errors through downstream tasks like object detection and mapping \cite{stutts2023lightweight}. Such errors can lead to severe consequences in safety-critical systems, including object misclassification or failures in collision avoidance. 

LiDAR power consumption is distributed across laser emission, scanning, signal processing, and data acquisition/control, with the total power defined as:
\[
P_{\text{total}} = P_{\text{laser}} + P_{\text{scan}} + P_{\text{signal}} + P_{\text{control}}.
\]
Here, the laser emitter's power (\(P_{\text{laser}}\)) depends on the energy per pulse (\(E_{\text{pulse}}\)), pulse repetition frequency (\(f_{\text{pulse}}\)), and laser efficiency (\(\eta_{\text{laser}}\)):
\[
P_{\text{laser}} = \frac{E_{\text{pulse}} \cdot f_{\text{pulse}}}{\eta_{\text{laser}}}.
\]
For scanning systems, power varies by design: mechanical systems consume 
\[
P_{\text{scan}} = \frac{V_{\text{motor}} \cdot I_{\text{motor}}}{\eta_{\text{motor}}},
\]
while MEMS and solid-state LiDAR systems require power for actuating MEMS mirrors or phase arrays. Signal processing power (\(P_{\text{signal}}\)) scales with computational complexity, and control power (\(P_{\text{control}}\)) involves tasks such as data handling via ADCs and microcontrollers.

These components result in several fundamental \textit{energy-accuracy-range trade-offs}. For example, for long-range sensing, the transmission energy (\(E_{\text{pulse}}\)) increases with the fourth power of range (\(R^4\)):
\[
E_{\text{pulse}} = \frac{P_r \cdot (4 \pi R^2)^2 \cdot \tau}{A_r \cdot \rho \cdot \eta},
\]
where \(P_r\) is the minimum received signal strength, \(A_r\) the receiver aperture area, \(\rho\) the target reflectivity, \(\eta\) the system efficiency, and \(\tau\) the laser pulse width. Similarly, finer range resolution (\(\Delta R = \frac{c \cdot \tau}{2}\)) requires shorter \(\tau\), increasing \(E_{\text{pulse}}\) and straining power and thermal budgets. Angular precision (\(\Delta \theta = \frac{\lambda}{D}\)) improves with larger apertures (\(D\)) or shorter wavelengths (\(\lambda\)), but these trade-offs impose challenges, including increased system footprint or eye-safety constraints. Higher precision also demands greater computational resources, with ADC power consumption scaling with sampling rate (\(f_s \geq \frac{c}{\Delta R}\)) and resolution (\(N\)):
\[
P_{\text{ADC}} = k \cdot \frac{c}{\Delta R} \cdot 2^N,
\]
where \(k\) is an ADC-specific constant. Signal processing tasks, such as FFTs, scale as \(O(N \log N)\) for large data sizes, further increasing energy demands.

Under these constraints, LiDAR systems at the edge become susceptible to various noise and degradation, typically categorized as \textit{data loss} and \textit{data corruption} \cite{nex2014uav,lu2019l3}. Data loss, arising from occlusions, reflective surfaces, or adverse weather conditions (e.g., rain, fog), leads to incomplete point clouds with missing spatial information. In contrast, data corruption stems from multi-path reflections, sensor vibrations, or atmospheric interference, producing erroneous points or clusters of spurious measurements. Both forms of noise degrade downstream tasks, where even minor errors can cascade into critical failures \cite{darabi2023starnet}. While preprocessing techniques such as statistical outlier removal or deep-learning-based denoising \cite{stutts2024mutual} can mitigate these issues to some extent, they often fall short in addressing the diverse and complex noise patterns encountered in real-world environments. This necessitates more advanced frameworks that can adaptively handle these challenges while maintaining efficiency and reliability at the edge.

\subsection{Deep Neural Networks for LiDAR Data Processing}  
DNNs have revolutionized 3D LiDAR data processing by addressing the irregular, unordered, and sparse nature of point clouds. PointNet \cite{bello2020deep} pioneered direct processing of raw point sets, avoiding information loss associated with voxelization. PointNet++ {\cite{qi2017pointnet++}} enhanced this approach with hierarchical feature extraction, enabling finer geometric representations for tasks like classification, segmentation, and object detection. Recent advancements, such as DGCNN {\cite{wang2019dynamic}}, utilize graph-based architectures to dynamically model local and global point interactions. Transformer-based models, including PointTransformer {\cite{zhao2021point}}, leverage self-attention mechanisms to capture long-range dependencies and focus on key regions, achieving state-of-the-art performance in tasks requiring both spatial precision and contextual awareness \cite{parente2024conformalized, darabi2024enhancing}.

Despite these advancements, DNNs remain vulnerable to noise and imperfections in LiDAR data. Environmental factors such as rain, fog, dust, and sensor issues like occlusions or multi-path reflections often result in corrupted or incomplete point clouds. These disruptions propagate errors through downstream tasks, degrading performance in safety-critical applications such as autonomous navigation {\cite{tayebati2024sense, darabi2024navigating}}, where noisy inputs can lead to misclassification or poor object localization, ultimately compromising system reliability. Models trained on clean or artificially augmented datasets frequently fail to generalize to the diverse and complex noise patterns encountered in real-world scenarios. Techniques like point dropout or jittering offer limited benefits, as they do not fully capture the intricacies of real-world noise. Even state-of-the-art architectures, such as PointTransformer and DGCNN, experience significant performance degradation under severe data corruption or sparsity {\cite{wicker2019robustness}}.

\subsection{Adversarial Curriculum Training and Meta-Learning}  
Robust learning strategies, such as data augmentation, noise injection, and adversarial training, have been extensively explored to enhance the resilience of DNNs. Among these, curriculum learning \cite{bengio2009curriculum} has emerged as a promising approach, structuring training with tasks or data samples of increasing difficulty to improve learning dynamics and generalization. Adversarial curriculum training (ACT) builds on this concept by progressively introducing higher levels of noise or perturbations during training. For LiDAR data, ACT systematically exposes the network to sensor noise and occlusions, starting with lightly corrupted inputs and gradually incorporating more complex noise patterns. This staged training enables the network to develop robust representations that generalize effectively to diverse and unpredictable noise distributions, making ACT particularly effective for safety-critical environments.

Meta-learning, or ``learning to learn," focuses on training algorithms to adapt rapidly to new conditions with minimal data \cite{finn2017model}. This approach is particularly relevant for LiDAR-based systems, where noise characteristics can vary significantly across environments and sensor configurations. Meta-learning techniques, such as Model-Agnostic Meta-Learning (MAML) \cite{finn2017model}, Reptile {\cite{jayathilaka2019enhancing}}, and Prototypical Networks {\cite{laenen2021episodes}}, have demonstrated success in few-shot learning and reinforcement learning and hold strong potential for 3D point cloud processing. In LiDAR applications, meta-learning can guide networks to acquire noise-resilient features that are easily fine-tuned to novel noise profiles, improving reliability in safety-critical scenarios.
\begin{figure}[t]
    \centering
    \includegraphics[width=0.99\linewidth]{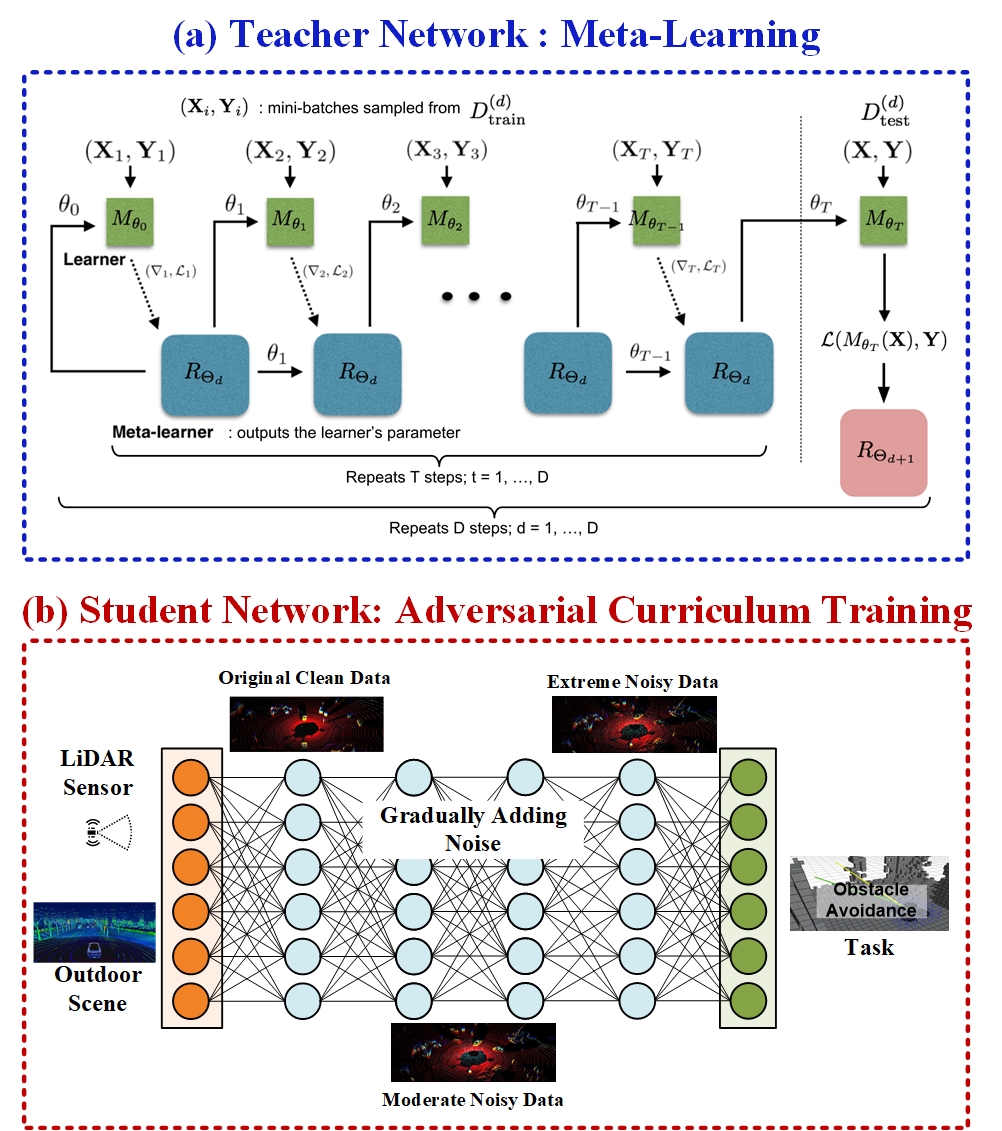}
    \caption{INTACT Overview: This figure shows overview of our proposed method. (a) Teacher Network: Meta-Learning phase trains a teacher network to identify saliency maps, guiding adversarial perturbation. (b) Student Network: Adversarial Curriculum Training phase trains a student network robust to noise by gradually increasing perturbation severity, guided by teacher saliency maps and a discriminator.}
    \label{fig:overview}
\end{figure}
\section{INTACT: Inducing Noise Tolerance through Adversarial Curriculum Training}  

INTACT employs a teacher-student network architecture, where the teacher leverages meta-learning to guide the student through adversarial curriculum training (ACT). This approach enhances robustness while maintaining computational efficiency, making it particularly suitable for resource-constrained robotic applications. Given a set of point clouds \( \mathcal{D} = \{P_1, P_2, \ldots, P_N\} \), where each point cloud \( P_i \in \mathbb{R}^{N_i \times 3} \), the framework aims to train a model \( M \) that is robust to sensor noise. When affected by noise \( \eta \in \mathbb{R}^{N_i \times 3} \), a point cloud transforms into \( P'_i = P_i + \eta \). INTACT addresses this challenge in two phases.  

\subsection{\underline{Phase I}: Meta-learning for the Teacher Network}  

In the first phase, a meta-learning framework trains the teacher network \( f_T \) to optimize its performance across a distribution of tasks. A task distribution \( \tau \sim p(\mathcal{T}) \) is defined, where each task involves classifying a subset of objects under varying noise conditions. During meta-iterations, a batch of tasks \( \{\mathcal{T}_1, \mathcal{T}_2, \ldots, \mathcal{T}_K\} \) is sampled, and the teacher is trained on each task. The meta-objective minimizes the expected loss across the task distribution:  
\begin{equation}
\theta_T^* = \arg \min_{\theta_T} \mathbb{E}_{\mathcal{T} \sim p(\mathcal{T})} [\mathcal{L}_{\mathcal{T}}(f_T(\theta_T))].
\label{eq:meta_objective}
\end{equation}  
This optimization enables the teacher to learn generalizable features, making it capable of adapting effectively to unseen noise distributions.  

Once meta-trained, the teacher generates gradients that identify the input points most critical to its predictions. For an input \( x \in \mathbb{R}^{N_i \times 3} \), the gradient with respect to the input is computed as:  
\begin{equation}
\nabla_x f_T(x) = \left( \frac{\partial f_T(x)}{\partial x_1}, \ldots, \frac{\partial f_T(x)}{\partial x_{N_i}} \right).
\end{equation}  
These gradients produce saliency maps, highlighting regions of the input point cloud that most influence the model’s output. The saliency maps serve as guidance for adversarial perturbations introduced in the next phase, ensuring that noise targets the most critical regions to enhance robustness.  

\subsection{\underline{Phase II}: Adversarial Training for the Student Network}

The second phase trains the student network \( f_S \) for deployment in real-world noisy environments. Using ACT, this phase progressively increases the severity of perturbations to stabilize the student's training and improve robustness. Key components of the process are detailed below:  

\vspace{4pt}
\noindent\textbf{(i) Student Network and Discriminator:}  
The student network \( f_S(\theta) \) learns robust representations by leveraging a discriminator \( f_{disc}(\phi) \), which generates challenging perturbations. These perturbations include noise injection or point removal, designed to test the student’s weaknesses. The training follows a min-max optimization framework:  
\begin{multline}
\theta_{opt} = \arg \min_{\theta} [L_{CE}(\theta) + \beta L_{Robust}(\theta, \phi) + \gamma L_{diff}(\theta) \\
+ \beta L_{curr\_Robust}(\theta, \phi) + \gamma L_{curr\_diff}(\theta)],
\label{eq:student_objective_min}
\end{multline}  
\begin{equation}
\phi_{opt} = \arg \max_{\phi} L_{Robust}(\theta, \phi).
\label{eq:discriminator_objective_max}
\end{equation}  
Here, \( L_{CE}(\theta) \) represents the cross-entropy loss measuring the student’s prediction error under unperturbed conditions. \( L_{Robust}(\theta, \phi) \) captures the vulnerability of the student to adversarial perturbations. The alignment loss \( L_{diff}(\theta) \) enforces gradient consistency between the teacher and student:  
\begin{equation}
L_{diff}(\theta) = ||\nabla_x f_S(x) - \nabla_x f_T(x)||^2.
\end{equation}  
Curriculum-based terms \( L_{curr\_Robust} \) and \( L_{curr\_diff} \) adapt perturbations as training progresses, ensuring gradual learning.  

\vspace{4pt}
\noindent\textbf{(ii) Perturbation Strategies:}  
Two adversarial perturbations are applied: 
\begin{itemize}
    \item \textit{Dropping Points}: A fraction \( \alpha \) of points is randomly removed, focusing on high-saliency regions identified by the teacher.
    \item \textit{Adding Noise}: Gaussian noise \( \delta \sim \mathcal{N}(0, \sigma^2) \) is injected into remaining points, with noise direction biased to maximize \( L_{Robust} \).
\end{itemize}

\vspace{4pt}
\noindent\textbf{(iii) Curriculum Design:}  
The curriculum progressively increases perturbation intensity. Initially, up to 90\% of saliency-sorted points are perturbed with modest Gaussian noise. Over iterations, the fraction of perturbed points decreases (e.g., from 90\% to 50\%), while the noise standard deviation \( \sigma_t \) increases incrementally:  
\[
\sigma_{t+1} = \sigma_t + \Delta \sigma.
\]  
This gradual progression stabilizes training, ensuring the student network gains robustness under broad conditions before specializing in handling severe perturbations.  

\vspace{4pt}
\noindent\textbf{(iv) Closed-Loop Training:}  
At each iteration \( t \), the framework updates the student and generates new adversarial samples:  

\begin{enumerate}
    \item The \textbf{teacher model} computes saliency scores for all data points in the current dataset \( D_t \), highlighting the most informative or influential samples for training.  

    \item The \textbf{discriminator} generates perturbations based on the extracted saliency scores, ensuring that modifications are guided by the most critical features of the data.  

    \item A \textbf{perturbed dataset} \( D_{t+1} \) is formed by applying the generated perturbations to the original data points, such that:
    \[
    D_{t+1} = \{P'_i = P_i + \eta_{t+1} \mid P_i \in D_t\}
    \]
    where \( \eta_{t+1} \) represents the noise introduced at this step.  

    \item The \textbf{student network} updates its parameters by minimizing the objective function defined in Eq.~\eqref{eq:student_objective_min}, incorporating the perturbed dataset to enhance robustness.  

    \item Over successive iterations, the \textbf{fraction of perturbed points gradually decreases}, while the \textbf{intensity of the applied noise increases}, ensuring a progressive adaptation to more challenging training conditions.  
\end{enumerate}

By the final stage, the student has been trained on highly perturbed samples focusing on critical regions, achieving robustness to structured and adversarial noise in real-world scenarios. The gradual progression of adversarial perturbations ensures efficient training while maintaining robustness in resource-constrained environments.  

\section{Results}

\subsection{Object Detection Results}

We evaluated the performance of three 3D object detection frameworks—PointPillar \cite{lang2019pointpillars}, SECOND \cite{yan2018second}, CenterFormer \cite{zhou2022centerformer}, and TED \cite{TED}—on the KITTI \cite{geiger2012we} and Argoverse \cite{chang2019argoverse} datasets. The evaluation considered various perturbation conditions, including 50\% point drop, Gaussian noise (\(\sigma = 0.1\)), and their combinations with Adversarial Curriculum Training (ACT) and Meta-Learning (ML). Detection accuracies were reported for three object classes: Car, Pedestrian, and Cyclist (KITTI), and Car, Pedestrian, and Bus (Argoverse). 

Tables \ref{tab:kitti_detection} and \ref{tab:argoverse_detection} summarize the results. Across all frameworks and datasets, the combination of ACT and ML consistently delivers the highest improvements in detection accuracy under both 50\% point drop and Gaussian noise conditions. For example, on the KITTI dataset with 50\% point drop, PointPillar’s Car class mAP improves from 56.1\% (baseline with 50\% drop) to 75.8\% with ACT+ML (i.e., INTACT). Similarly, on the Argoverse dataset with Gaussian noise, SECOND’s Bus class mAP increases from 38.6\% (baseline with noise) to 69.8\% with INTACT. These results demonstrate the effectiveness of ACT and ML in significantly enhancing model robustness across various object classes, datasets, and architectures.

The improvements achieved by ACT and ML are consistent across all object classes (Car, Pedestrian, Cyclist/Bus) and frameworks, underscoring the broad applicability of these techniques. Notably, improvements are more pronounced for the Car and Cyclist/Bus classes compared to Pedestrian detection, likely due to structural and geometric differences that affect how noise impacts detectability. Among the frameworks, SECOND exhibits the greatest improvement when ACT and ML are applied, particularly under noisy conditions, suggesting it benefits the most from advanced training techniques compared to PointPillar and CenterFormer. 

While CenterFormer demonstrates high baseline robustness—particularly in Pedestrian detection—it still shows measurable gains with ACT and ML. This highlights that even robust models can benefit from INTACT. These findings illustrate INTACT's ability to improve robustness across a wide range of detection models and perturbation scenarios, further advancing the state-of-the-art in 3D object detection.

\begin{table*}[t]
\centering
\caption{Object detection performance on the KITTI dataset under baseline, 50\% point drop, and Gaussian noise (\(\sigma = 0.1\)) conditions. Results are reported for various architectures and object classes (Car, Pedestrian, Cyclist). ACT refers to Adversarial Curriculum Training, and ML refers to Meta-Learning.}
\label{tab:kitti_detection}
\begin{tabular}{l||ccc|ccc|ccc|ccc}
\hline
\multirow{2}{*}{\textbf{Object Class}} & \multicolumn{3}{c|}{\textbf{PointPillar} {\cite{lang2019pointpillars}}} & \multicolumn{3}{c|}{\textbf{SECOND} {\cite{yan2018second}}} & \multicolumn{3}{c|}{\textbf{CenterFormer} {\cite{zhou2022centerformer}}} & \multicolumn{3}{c}{\textbf{TED} \cite{TED}} \\
& \textbf{Car} & \textbf{Pedestrian} & \textbf{Cyclist} & \textbf{Car} & \textbf{Pedestrian} & \textbf{Cyclist} & \textbf{Car} & \textbf{Pedestrian} & \textbf{Cyclist} & \textbf{Car} & \textbf{Pedestrian} & \textbf{Cyclist} \\
\hline \hline
Baseline & 77.3 & 52.3 & 62.7 & 78.6 & 53.0 & 67.1 & 75.0 & 78.0 & 73.8 & 78.1 & 81.4 & 76.3 \\
50\% Drop & 56.1 & 34.2 & 51.6 & 65.3 & 37.1 & 54.2 & 70.1 & 72.8 & 65.3 & 72.5 & 73.9 & 66.1 \\
50\% Drop + ACT & 68.4 & 45.5 & 57.2 & 71.5 & 46.1 & 62.8 & 73.5 & 75.1 & 70.7 & 75.3 & 77.2 & 70.2 \\
50\% Drop + INTACT & \textbf{75.8} & \textbf{51.0} & \textbf{60.7} & \textbf{77.2} & \textbf{51.8} & \textbf{66.8} & \textbf{74.3} & \textbf{76.9} & \textbf{71.9} & \textbf{77.8} & \textbf{79.5} & \textbf{72.6} \\
\hline \hline
Noise & 50.9 & 35.4 & 50.1 & 57.3 & 42.1 & 20.7 & 55.8 & 39.2 & 50.7 & 56.7 & 40.5 & 51.9 \\
Noise + ACT & 67.2 & 47.1 & 57.3 & 68.7 & 47.0 & 64.7 & 70.2 & 49.8 & 65.5 & 71.7 & 51.2 & 66.7 \\
Noise + INTACT & \textbf{73.0} & \textbf{50.7} & \textbf{59.9} & \textbf{76.5} & \textbf{50.5} & \textbf{66.2} & \textbf{77.1} & \textbf{52.9} & \textbf{67.3} & \textbf{78.6} & \textbf{54.1} & \textbf{68.4} \\
\hline
\end{tabular}
\end{table*}

\begin{table*}[t]
\centering
\caption{Object detection performance on the Argoverse dataset under baseline, 50\% point drop, and Gaussian noise (\(\sigma = 0.1\)) conditions. Results are reported for various architectures and object classes (Car, Pedestrian, Bus). ACT refers to Adversarial Curriculum Training, and ML refers to Meta-Learning.}
\label{tab:argoverse_detection}
\begin{tabular}{l||ccc|ccc|ccc|ccc}
\hline
\multirow{2}{*}{\textbf{Object Class}} & \multicolumn{3}{c|}{\textbf{PointPillar} {\cite{lang2019pointpillars}}} & \multicolumn{3}{c|}{\textbf{SECOND} {\cite{yan2018second}}} & \multicolumn{3}{c|}{\textbf{CenterFormer} {\cite{zhou2022centerformer}}} & \multicolumn{3}{c}{\textbf{TED} \cite{TED}} \\
& \textbf{Car} & \textbf{Pedestrian} & \textbf{Bus} & \textbf{Car} & \textbf{Pedestrian} & \textbf{Bus} & \textbf{Car} & \textbf{Pedestrian} & \textbf{Bus} & \textbf{Car} & \textbf{Pedestrian} & \textbf{Bus} \\
\hline
\hline
Baseline & 70.5 & 59.9 & 34.4 & 72.1 & 58.3 & 40.5 & 71.7 & 69.1 & 51.9 & 73.3 & 71.5 & 53.8 \\
50\% Drop & 57.3 & 42.1 & 20.7 & 56.8 & 44.5 & 27.4 & 57.2 & 58.5 & 39.5 & 58.6 & 59.1 & 40.2 \\
50\% Drop + ACT & 64.9 & 53.3 & 27.5 & 67.1 & 52.0 & 34.8 & 66.1 & 66.3 & 47.1 & 67.8 & 68.4 & 48.3 \\
50\% Drop + INTACT & \textbf{69.3} & \textbf{58.2} & \textbf{31.0} & \textbf{71.2} & \textbf{56.1} & \textbf{38.7} & \textbf{68.9} & \textbf{68.4} & \textbf{50.1} & \textbf{72.6} & \textbf{70.2} & \textbf{52.1} \\
\hline
\hline
Noise & 51.6 & 39.7 & 17.4 & 54.1 & 39.7 & 23.4 & 55.1 & 54.0 & 37.7 & 56.2 & 56.1 & 38.9 \\
Noise + ACT & 62.6 & 51.9 & 25.5 & 65.0 & 55.8 & 36.5 & 66.9 & 64.5 & 45.9 & 68.4 & 66.0 & 47.5 \\
Noise + INTACT & \textbf{68.7} & \textbf{57.5} & \textbf{32.3} & \textbf{70.1} & \textbf{57.2} & \textbf{38.6} & \textbf{69.8} & \textbf{66.9} & \textbf{50.2} & \textbf{71.2} & \textbf{68.3} & \textbf{51.4} \\
\hline
\end{tabular}
\end{table*}

\subsection{Classification Results}

Table \ref{tab:classification} reports the classification accuracy of three point cloud classification architectures—PointNet {\cite{qi2017pointnet}}, PointNet++ {\cite{qi2017pointnet++}}, and Point Transformer {\cite{zhao2021point}}—on the ModelNet40 dataset under various input perturbation conditions. The evaluation considers baseline performance and the impact of 50\% point drop, Gaussian noise (\(\sigma = 0.1\)), and their combinations with Adversarial Curriculum Training (ACT) and Meta-Learning (ML). Among the baseline models, Point Transformer achieves the highest accuracy (93.8\%), followed by PointNet++ (92.5\%) and PointNet (90.4\%), reflecting the architectural advancements of Point Transformer in capturing geometric features. However, all models exhibit significant performance degradation under perturbations. For instance, PointNet’s accuracy drops from 90.4\% in the baseline to 58.8\% with 50\% point drop, highlighting its sensitivity to missing data.

The introduction of ACT significantly improves robustness across all models under noise. For example, PointNet++ with ACT achieves 89.9\% accuracy under 50\% point drop, a notable improvement over the baseline's 86.4\% in the same condition. Adding Meta-Learning (ACT+ML, i.e., INTACT) further enhances performance, yielding the best recovery under perturbations. For instance, Point Transformer with INTACT achieves 92.7\% accuracy under 50\% point drop, nearly matching its baseline performance of 93.8\%. Point Transformer demonstrates the highest robustness and recovery across all perturbation types, followed by PointNet++ and PointNet. This resilience can be attributed to its attention mechanisms, which enhance its ability to handle point cloud perturbations. Nonetheless, the INTACT framework consistently improves the robustness of all architectures, underscoring its general applicability and effectiveness in handling noisy inputs.

\begin{table*}[t]
\centering
\caption{Classification performance on ModelNet40: Overall accuracy (\%) under two perturbation types (50\% point drop and Gaussian noise with \(\sigma = 0.1\)). ACT refers to Adversarial Curriculum Training, and ML refers to Meta-Learning. Results are reported for five architectures.}
\label{tab:classification}
\begin{tabular}{l||ccccc}
\hline
\textbf{Condition} & \textbf{PointNet} {\cite{qi2017pointnet}} & \textbf{PointNet++} {\cite{qi2017pointnet++}} & \textbf{DGCNN} {\cite{wang2019dynamic}} & \textbf{PointConv} \cite{wu2019pointconv} & \textbf{Point Transformer} {\cite{zhao2021point}} \\
\hline\hline
Baseline & 90.4 & 92.5 & 92.2 & 93.0 & 93.8 \\
50\% Point Drop & 58.8 & 86.4 & 84.1 & 85.7 & 87.2 \\
50\% Point Drop + ACT & 87.3 & 89.9 & 89.1 & 90.4 & 91.6 \\
50\% Point Drop + INTACT & \textbf{89.7} & \textbf{90.2} & \textbf{90.9} & \textbf{91.2} & \textbf{92.7} \\
\hline\hline
Gaussian Noise (\(\sigma = 0.1\)) & 65.4 & 62.6 & 75.3 & 79.6 & 81.8 \\
Gaussian Noise + ACT & 77.5 & 77.8 & 85.2 & 88.1 & 90.7 \\
Gaussian Noise + INTACT & \textbf{82.6} & \textbf{83.2} & \textbf{87.6} & \textbf{90.2} & \textbf{92.1} \\
\hline
\end{tabular}
\end{table*}

\subsection{Object Tracking Results}
Table \ref{tab:tracking} summarizes the object tracking performance of the YOLOv8n model on the KITTI and Argoverse tracking datasets. We evaluate performance using two key metrics: Multiple Object Tracking Accuracy (MOTA) and mean Average Precision (mAP). These metrics measure the accuracy and precision of tracking under baseline and noisy conditions, including 50\% point drop and Gaussian noise. We also analyze the impact of integrating Adversarial Curriculum Training (ACT) and Meta-Learning (ML) on performance recovery.

The baseline tracking performance on the KITTI dataset is high, with MOTA and mAP values of 78.5\% and 72.9\%, respectively. However, noisy conditions, such as a 50\% point drop, result in substantial performance degradation, reducing MOTA to 64.1\% and mAP to 59.8\%. Similarly, Gaussian noise lowers MOTA to 52.5\% and mAP to 49.3\%, underscoring the sensitivity of object tracking to input perturbations.

Applying ACT alone significantly improves tracking performance under noisy conditions. For instance, under a 50\% point drop, MOTA increases from 64.1\% to 70.2\%, and mAP improves from 59.8\% to 65.7\%. The addition of Meta-Learning (INTACT) further enhances robustness, bringing performance closer to baseline levels. Under a 50\% point drop, INTACT achieves a MOTA of 75.1\% and an mAP of 69.8\%, representing a substantial recovery compared to ACT alone. Similar improvements are observed for Gaussian noise, where INTACT achieves a MOTA of 73.7\% and an mAP of 70.9\%.

These results highlight the effectiveness of INTACT in improving the robustness of DNNs for object tracking, even under significant noise and point cloud degradation. By integrating ACT and Meta-Learning, the framework demonstrates its ability to recover performance across various metrics and conditions, making it highly applicable for safety-critical tasks in dynamic and noisy environments.

\begin{table}[t]
\centering
\caption{Object tracking performance using the YOLOv8n model. MOTA represents Multiple Object Tracking Accuracy, and mAP denotes mean Average Precision. Results are reported for the KITTI {\cite{geiger2012we}} and Argoverse {\cite{chang2019argoverse}} datasets under various noise and training conditions.}
\label{tab:tracking}
\begin{tabular}{l||cc|cc}
\hline
\textbf{Condition} & \multicolumn{2}{c|}{\textbf{KITTI}} & \multicolumn{2}{c}{\textbf{Argoverse}} \\
 & \textbf{MOTA} & \textbf{mAP} & \textbf{MOTA} & \textbf{mAP} \\
\hline\hline
Baseline & 78.5 & 72.9 & 51.5 & 33.8 \\
50\% Point Drop & 64.1 & 59.8 & 37.2 & 20.4 \\
50\% Point Drop + ACT & 70.2 & 65.7 & 44.9 & 27.2 \\
50\% Point Drop + INTACT & \textbf{75.1} & \textbf{69.8} & \textbf{48.9} & \textbf{31.9} \\
\hline\hline
Gaussian Noise & 52.5 & 49.3 & 32.5 & 19.3 \\
Gaussian Noise + ACT & 65.8 & 62.7 & 43.7 & 26.5 \\
Gaussian Noise + INTACT & \textbf{73.7} & \textbf{70.9} & \textbf{49.1} & \textbf{30.7} \\
\hline
\end{tabular}
\end{table}

\section{Conclusion}
We introduced INTACT, a two-phase framework that combines meta-learning with curriculum-based adversarial training to enhance the robustness of 3D point cloud networks. By leveraging saliency maps from the meta-trained teacher, perturbations are directed to critical regions, enabling the student to learn robust features while adapting to adversarial noise. This approach improves resilience to common challenges such as point dropping and Gaussian noise while maintaining computational efficiency, making it well-suited for resource-constrained platforms. Experiments across object detection, classification, and tracking tasks demonstrate that INTACT consistently outperforms baseline models and standard adversarial training under various noise conditions. The framework’s ability to generalize across tasks and architectures underscores its versatility and practicality for real-world applications.

Future directions include incorporating diverse perturbation types, such as global transformations and realistic sensor noise models, to address more complex noise patterns. Hierarchical saliency maps could provide finer guidance for perturbations, while real-time adaptation of noise parameters based on dynamic conditions could further enhance robustness.

\vspace{2pt}
\noindent\textbf{Acknowledgment:} This work was partially supported by COGNISENSE, one of seven centers in JUMP 2.0, a Semiconductor Research Corporation (SRC) program sponsored by DARPA and NSF Awards \#2046435.

\vspace{2pt}
\noindent\textbf{Use of AI:}
ChatGPT4o was used to polish the text and fix grammar, ensuring clarity, consistency, and a professional tone while preserving the original meaning.

\bibliographystyle{IEEEtran}
\bibliography{main}

\end{document}